\documentclass[11pt]{article}
\usepackage[]{acl}

\usepackage{times}
\usepackage{latexsym}
\usepackage[T1]{fontenc}
\usepackage[utf8]{inputenc}
\usepackage{microtype}
\usepackage{inconsolata}
\usepackage{graphicx}
\usepackage{booktabs}
\usepackage{multirow}
\usepackage{colortbl}
\usepackage{xcolor}
\usepackage{amsmath,amssymb,amsthm}
\usepackage{enumitem}
\usepackage{tikz}
\usetikzlibrary{positioning,shapes.geometric,arrows.meta,calc,fit,
                backgrounds,decorations.pathreplacing}
\usepackage{xspace}
\usepackage[most]{tcolorbox}

\definecolor{colA}{HTML}{4E79A7}
\definecolor{colB}{HTML}{F28E2B}
\definecolor{colC}{HTML}{E15759}
\definecolor{colD}{HTML}{59A14F}
\definecolor{colE}{HTML}{76B7B2}
\definecolor{colF}{HTML}{EDC948}
\definecolor{colG}{HTML}{B07AA1}
\definecolor{lightbg}{HTML}{F4F6F8}
\definecolor{midgr}{HTML}{CCCED2}
\definecolor{darkgr}{HTML}{5C6470}

\newtheorem{proposition}{Proposition}[section]
\newtheorem{definition}{Definition}[section]
\newtheorem{remark}{Remark}

\newtcolorbox{promptbox}[1][]{enhanced,colback=lightbg,colframe=darkgr,
  boxrule=0.4pt,arc=2pt,left=4pt,right=4pt,top=2pt,bottom=2pt,
  fonttitle=\small\bfseries,#1}

\newcommand{\adarubric}{\textsc{AdaRubric}\xspace}
\newcommand{\rowhl}{\rowcolor{colA!10}}
\newcommand{\eg}{e.g.,\xspace}
\newcommand{\ie}{i.e.,\xspace}

\title{\textsc{AdaRubric}: Task-Adaptive Rubrics for\\
Reliable LLM Agent Evaluation and Reward Learning}

\author{Liang Ding \\
  The University of Sydney \\
  \texttt{liangding.liam@gmail.com}
}

\begin{document}
\maketitle

\begin{abstract}
Evaluating LLM agent trajectories is fundamentally \emph{task-specific}:
a code-debugging agent should be judged on \emph{Correctness} and
\emph{Error Handling}, not on \emph{Fluency} or \emph{Safety}.
Yet the dominant paradigm---LLM-as-Judge with a fixed rubric---applies the
same static dimensions regardless of task, producing systematic mis-evaluation.
We present \textbf{\adarubric}, a framework that
(i)~\emph{adaptively generates} task-specific evaluation rubrics from task descriptions via LLM,
(ii)~evaluates agent trajectories step-by-step with confidence-weighted,
per-dimension scoring, and
(iii)~produces dense reward signals for preference learning.
Three composable filtering strategies, including the novel
\emph{DimensionAwareFilter} that provably prevents dimension-level quality
masking, yield high-quality DPO preference pairs.
On WebArena, ToolBench, and AgentBench, \adarubric achieves
\textbf{Pearson $r$\,=\,0.79} human correlation ($+0.15$ over the strongest baseline),
with strong reliability (Krippendorff's $\alpha$\,=\,0.83).
DPO models trained on \adarubric-generated pairs improve task success by
\textbf{$+6.8$--$+8.5$\%} over the best baseline.
\adarubric also generalises zero-shot to unseen domains (SWE-bench)
and extends to multimodal agents (VisualWebArena, OSWorld) without modification. Our code is available at: \url{github.com/alphadl/AdaRubrics}
\end{abstract}

\section{Introduction}
\label{sec:intro}

LLM now automates complex multi-step tasks across web automation
\citep{zhou2024webarena}, API orchestration \citep{qin2024toolbench},
software engineering \citep{jimenez2024swebench},
and multimodal environments \citep{koh2024visualwebarena,xie2024osworld}.
As agents scale, reliable trajectory evaluation becomes the cornerstone of
safety, alignment, and iterative improvement.
Yet studies show that a substantial fraction of agent trajectories produced by
frontier models on representative benchmarks contain steps of uncertain quality
\citep{liu2024agentbench,pan2024autonomous}, making undifferentiated quality
judgments insufficient for training or deployment.
A key question for the knowledge-in-LMs community is:
\emph{how should an LLM evaluator leverage its knowledge to generate
task-appropriate evaluation criteria, rather than applying a fixed,
one-size-fits-all rubric?}

\paragraph{The static-rubric bottleneck.}
Two dominant evaluation paradigms fail for complex agents.
\emph{Reference-based metrics} (ROUGE-L, BERTScore) measure surface overlap and
are blind to goal-directed reasoning.
\emph{LLM-as-Judge} \citep{zheng2023judging,liu2023geval} applies fixed
dimensions---\emph{Helpfulness}, \emph{Fluency}, \emph{Safety}---regardless of task.
These dimensions were designed for chat assistants, not goal-directed agents
operating through tool use and multi-step planning.

\paragraph{Motivating example.}
A ToolBench API-chaining task has meaningful quality dimensions of
\emph{API Selection Accuracy}, \emph{Parameter Correctness}, and
\emph{Error Recovery}---none appearing in a standard helpfulness rubric.
Conversely, a fluency-focused rubric penalises agents that correctly call APIs
but produce compact, machine-readable output.
Static rubrics introduce systematic bias and reward irrelevant stylistic
properties over task success.

\begin{figure}[t]
\centering
\resizebox{\columnwidth}{!}{%
\begin{tikzpicture}[
  font=\small,
  taskbox/.style={draw=darkgr!60,fill=lightbg,rounded corners=3pt,
                  minimum width=2.0cm,minimum height=0.50cm,
                  align=center,inner sep=2pt,font=\footnotesize},
  rubbox/.style={draw,rounded corners=2pt,minimum width=1.6cm,
                 minimum height=0.38cm,align=center,
                 inner sep=2pt,font=\scriptsize},
  static/.style={rubbox,fill=colC!10,draw=colC!45},
  adapt/.style={rubbox,fill=colA!10,draw=colA!45},
  arrow/.style={-Stealth,thick,darkgr!80},
  slbl/.style={font=\footnotesize\bfseries},
]
  \node[taskbox] (tA) at (-3.8,1.2) {Code Debug};
  \node[taskbox] (tB) at (-3.8,0.0) {Web Search};

  \node[draw=colC!55,fill=colC!5,rounded corners=3pt,
        minimum width=1.7cm,minimum height=1.4cm,
        label={[font=\scriptsize\bfseries,colC!80]above:Static Rubric}]
        (fbox) at (-1.3,0.6) {};
  \node[static,minimum width=1.5cm] at (-1.3,1.0) {Helpfulness};
  \node[static,minimum width=1.5cm] at (-1.3,0.6) {Fluency};
  \node[static,minimum width=1.5cm] at (-1.3,0.2) {Safety};

  \draw[colC!65, thick] (tA.east) -| ++(0.35,-0.6) coordinate (joinLeft);
  \draw[colC!65, thick] (tB.east) -| (joinLeft);
  \draw[arrow,colC!65] (joinLeft) -- (fbox.west);

  \node[slbl,colC] at (0.35,1.2) {\scriptsize$r{=}0.47$};
  \node[slbl,colC] at (0.35,0.0) {\scriptsize$r{=}0.44$};

  \node[font=\scriptsize\bfseries,colC!85] at (-1.8,-0.7)
        {(a) Static Eval.};

  \node[taskbox] (tC) at (2.3,1.3) {Code Debug};
  \node[taskbox] (tD) at (2.3,-0.1) {Web Search};

  \node[draw=colA!55,fill=colA!5,rounded corners=3pt,
        minimum width=1.7cm,minimum height=1.1cm,
        label={[font=\tiny\bfseries,colA!80]above:Ada (Code)}]
        (rA) at (4.9,1.3) {};
  \node[adapt,minimum width=1.5cm] at (4.9,1.6) {\tiny Correctness};
  \node[adapt,minimum width=1.5cm] at (4.9,1.3) {\tiny Err. Handle};
  \node[adapt,minimum width=1.5cm] at (4.9,1.0) {\tiny Efficiency};

  \node[draw=colA!55,fill=colA!5,rounded corners=3pt,
        minimum width=1.7cm,minimum height=1.1cm,
        label={[font=\tiny\bfseries,colA!80]above:Ada (Search)}]
        (rB) at (4.9,-0.1) {};
  \node[adapt,minimum width=1.5cm] at (4.9,0.2) {\tiny Coverage};
  \node[adapt,minimum width=1.5cm] at (4.9,-0.1) {\tiny Tool Acc.};
  \node[adapt,minimum width=1.5cm] at (4.9,-0.4) {\tiny Synthesis};

  \draw[arrow,colA!65] (tC.east) -- (rA.west);
  \draw[arrow,colA!65] (tD.east) -- (rB.west);

  \node[slbl,colD] at (6.3,1.3) {\scriptsize$r{=}0.79$};
  \node[slbl,colD] at (6.3,-0.1) {\scriptsize$r{=}0.74$};

  \node[font=\scriptsize\bfseries,colA!85] at (3.4,-0.7)
        {(b) \adarubric};

  \draw[dashed,thick,darkgr!55] (1.3,-0.9) -- (1.3,2.1);
\end{tikzpicture}%
}
\vspace{-6pt}
\caption{\textbf{Static evaluation vs.\ \adarubric.}
Static LLM-as-Judge applies identical dimensions to all tasks ($r{\approx}0.46$).
\adarubric synthesises task-specific rubrics ($r{\approx}0.77$).}
\label{fig:motivation}
\vspace{-8pt}
\end{figure}

\paragraph{AdaRubric.}
We propose \adarubric, built on the insight:
\emph{evaluation dimensions should be a function of the task, not a fixed
property of the evaluator}.
Given a task description $T$, \adarubric generates a
\emph{Dynamic Rubric} $\mathcal{R}(T)$ comprising $N$ task-specific,
orthogonal evaluation dimensions with calibrated 5-point scoring criteria.
This approach directly leverages the LLM's parametric knowledge about
task structure, success criteria, and domain conventions to produce
evaluations aligned with human expert judgment.
Figure~\ref{fig:motivation} shows the contrast between static and adaptive
evaluation.

\paragraph{Contributions.}
\begin{enumerate}[leftmargin=*,topsep=2pt,itemsep=1pt]
\item \textbf{Adaptive rubric generation} (\S\ref{sec:stage1}):
  task-specific, orthogonal evaluation dimensions with calibrated scoring criteria,
  generated from task descriptions via LLMs' parametric knowledge.
\item \textbf{Multi-dimensional dense rewards} (\S\ref{sec:stage2}):
  confidence-weighted, per-step, per-dimension scoring for post-training process, e.g., RL/DPO.
\item \textbf{Reliability quantification} (\S\ref{sec:reliability}):
  Krippendorff's $\alpha$ provides a principled deployment
  criterion ($\alpha \geq 0.80$) for LLM-based evaluators.
\item \textbf{End-to-end evaluation-to-training pipeline} (\S\ref{sec:stage3}):
  composable filters including the novel \emph{DimensionAwareFilter}
  that provably prevents per-dimension quality masking.
\end{enumerate}

\section{Related Work}
\label{sec:related}

\paragraph{LLM-as-Judge evaluation.}
\citet{zheng2023judging} established MT-Bench and Chatbot Arena using pairwise comparison with fixed dimensions.
G-Eval \citep{liu2023geval} employs GPT-4 with explicit criteria for NLG.
Prometheus \citep{kim2024prometheus} fine-tunes a 13B judge requiring labelled
training data per task domain.
FLASK \citep{ye2024flask} decomposes quality into fine-grained skill sets;
JudgeLM \citep{zhu2023judgelm} trains judges from large (question, answer,
judgment) corpora.
RewardBench \citep{lambert2024rewardbench} provides a systematic benchmark
for comparing reward models across task types, sharpening the need for
task-sensitive evaluation criteria.
\citet{lu2023toward} decompose NLG evaluation into major and minor error axes,
an early instantiation of structured, rubric-like scoring that motivates our
task-adaptive dimension design.
Yet all these methods rely on fixed or pre-trained dimensions; \adarubric
closes this gap by generating criteria dynamically from the task at hand,
without manual design or fine-tuning.

\paragraph{LLM agent evaluation.}
WebArena \citep{zhou2024webarena}, ToolBench \citep{qin2024toolbench},
AgentBench \citep{liu2024agentbench}, and SWE-bench \citep{jimenez2024swebench}
provide task-specific success signals.
\citet{pan2024autonomous} proposes autonomous evaluation by output comparison.
\citet{lu2025runaway} study agent trajectory quality from a different angle,
showing that early-exit strategies reduce redundant steps in embodied agents
and introducing efficiency and progress metrics that are natural reward components.
These signals are binary or coarse-grained, non-transferable across tasks,
and offer no per-step reward signal suitable for RL training---a limitation
\adarubric addresses directly.

\paragraph{Reward signals and RLHF/DPO.}
RLHF \citep{christiano2017deep,ziegler2019fine,stiennon2020learning,ouyang2022training}
trains scalar reward models; DPO \citep{rafailov2023direct} eliminates the
explicit reward model entirely.
T\"{u}lu \citep{wang2023far,ivison2023camels} shows that preference data quality
critically determines RLHF effectiveness.
Process reward models \citep{lightman2024lets,cobbe2021training} assign
step-level credits for math reasoning; self-rewarding LMs \citep{yuan2024self}
close the loop.
Most recently, DR~Tulu \citep{shao2025drtulu} introduces
\emph{Reinforcement Learning with Evolving Rubrics} (RLER), where
instance-specific, search-grounded rubrics co-evolve with the policy
during RL training for long-form deep research tasks.
\adarubric differs in three key respects:
(i)~rubrics are generated \emph{per task type} (not per instance) and cached,
making evaluation $>$95\% cheaper;
(ii)~rubrics derive from the LLM's \emph{parametric knowledge} rather than
external retrieval;
(iii)~the focus is \emph{agent trajectory evaluation and reward synthesis}
rather than long-form QA training.
The two approaches are complementary: \adarubric's task-adaptive rubrics
could serve as the initial rubric framework that RLER then evolves
online during training.

\paragraph{AgentHER and trajectory augmentation.}
One concurrent work \citep{ding2026agentHER} relabels failed agent
trajectories via hindsight experience replay, focusing on \emph{data
augmentation}.
\adarubric focuses on \emph{principled evaluation and reward synthesis},
and the two approaches are naturally complementary.

\paragraph{Inter-rater reliability.}
Krippendorff's $\alpha$ \citep{krippendorff2011computing} and Fleiss'
$\kappa$ \citep{fleiss1971measuring} quantify human annotation agreement.
We apply these metrics to quantify \emph{LLM evaluator consistency} across
evaluation runs, providing a principled criterion for deployment.

\section{The \adarubric Framework}
\label{sec:method}

\subsection{Problem Formulation}
\label{sec:formulation}

A \textbf{task} $T = (i, d, c, E)$ comprises instruction $i$, domain $d$,
context $c$, and expected tools/modalities $E$.
An \textbf{agent trajectory} $\tau = \{(t_k, a_k, o_k)\}_{k=1}^K$ consists
of $K$ steps of (thought, action, observation).
Static evaluation maps $\tau$ to a scalar; \adarubric produces structured,
task-conditioned evaluations:
\begin{equation}
  f_\text{ada}(\tau;\,\mathcal{R}(T))
  \to \bigl\{(s_{k,j},\, c_{k,j})\bigr\}_{k=1,\,j=1}^{K,\,N},
\end{equation}
where $s_{k,j} \in \{1,\ldots,5\}$ and $c_{k,j} \in [0,1]$ is the confidence
(step relevance to dimension $j$).

\begin{definition}[Task-Adaptive Rubric]
\label{def:rubric}
$\mathcal{R}(T) = \{(d_j, w_j, \Gamma_j)\}_{j=1}^N$, where $d_j$ is a
dimension name, $w_j > 0$ with $\sum_j w_j = 1$, and
$\Gamma_j = (\gamma_1^j,\ldots,\gamma_5^j)$ are verbalized scoring criteria.
A valid rubric satisfies:
(i)~\emph{Task-relevance}: $d_j$ derived from $T$'s success criteria;
(ii)~\emph{Orthogonality}: dimensions are semantically non-overlapping;
(iii)~\emph{Completeness}: $\bigcup_j d_j$ covers $T$'s key success aspects;
(iv)~\emph{Calibration}: $\gamma_3^j =$\,``acceptable'',
$\gamma_1^j =$\,``broken'', $\gamma_5^j =$\,``exemplary''.
\end{definition}

\adarubric operates through three stages (Figure~\ref{fig:pipeline}) followed
by reward synthesis.

\begin{figure}[t]
\centering
\resizebox{\columnwidth}{!}{%
\begin{tikzpicture}[
  font=\small,
  stage/.style={draw,rounded corners=4pt,minimum width=2.0cm,
                minimum height=0.85cm,align=center,inner sep=3pt},
  iobox/.style={draw,rounded corners=3pt,fill=lightbg,
                minimum width=1.6cm,minimum height=0.42cm,
                align=center,font=\scriptsize,inner sep=2pt},
  arrow/.style={-Stealth,thick,darkgr!80},
  badge/.style={draw,circle,font=\bfseries\scriptsize,
                inner sep=1pt,minimum size=11pt},
]
  \node[iobox,fill=colA!10,draw=colA!50] (task) at (-3.2,0.4)
    {\textbf{Task} $T$};
  \node[iobox,fill=colA!10,draw=colA!50] (trajs) at (-3.2,-0.4)
    {\textbf{Trajs} $\{\tau_i\}$};

  \node[stage,fill=colA!12,draw=colA] (s1) at (-0.9,0.4) {
    \textbf{S1}\\{\scriptsize Rubric Gen.}};
  \node[badge,fill=colA!20,draw=colA] at (-0.9,0.95) {\color{colA}1};

  \node[stage,fill=colB!12,draw=colB] (s2) at (1.5,-0.4) {
    \textbf{S2}\\{\scriptsize Evaluator}};
  \node[badge,fill=colB!20,draw=colB] at (1.5,0.15) {\color{colB}2};

  \node[stage,fill=colD!12,draw=colD] (s3) at (3.9,-0.4) {
    \textbf{S3}\\{\scriptsize Filter}};
  \node[badge,fill=colD!20,draw=colD] at (3.9,0.15) {\color{colD}3};

  \node[iobox,fill=colE!12,draw=colE!60] (dpo) at (5.8,-0.4)
    {\textbf{DPO Pairs}};

  \draw[arrow] (task.east) -- (s1.west);
  \draw[arrow] (s1.east) -- ++(0.25,0) |- (s2.west);
  \draw[arrow] (trajs.east) -- ++(0.25,0) |- (s2.west);
  \draw[arrow] (s2.east) -- (s3.west);
  \draw[arrow] (s3.east) -- (dpo.west);
\end{tikzpicture}%
}
\vspace{-4pt}
\caption{\textbf{\adarubric pipeline.}
Stage~1 synthesises a task-adaptive rubric.
Stage~2 evaluates trajectories per-step $\times$ per-dimension.
Stage~3 applies composable filters and generates DPO pairs.}
\label{fig:pipeline}
\vspace{-8pt}
\end{figure}

\begin{figure}[t]
\begin{promptbox}[title=Algorithm 1: \adarubric Evaluation Pipeline]
\textbf{Require:} Task $T$, trajectories $\{\tau_i\}$, LLM $\mathcal{M}$,
params $N, \lambda, \delta_\text{min}$\\
\textbf{Ensure:} DPO pairs $\mathcal{P}$, scores $\{S(\tau_i)\}$
\vspace{3pt}
\hrule
\vspace{3pt}
\textbf{1. Rubric Generation}\\
\quad $\mathcal{R}(T) \leftarrow \mathcal{M}(\text{RUBRIC\_PROMPT}(T))$
  \hfill \textit{// adaptive, cached}\\
\quad Validate $\mathcal{R}$; retry once on failure

\vspace{2pt}
\textbf{2. Trajectory Evaluation} \textbf{for each} $\tau_i$:\\
\quad \textbf{for} $k=1\ldots K$, $j=1\ldots N$ \textbf{do}\\
\quad\quad $(s_{k,j}, c_{k,j}) \leftarrow \mathcal{M}(\text{EVAL\_PROMPT}(t_k, a_k, o_k, d_j, \Gamma_j))$\\
\quad Aggregate: $\bar{s}_j \leftarrow \textsc{WM}(s_{\cdot,j}, c_{\cdot,j}, \lambda)$\\
\quad $S(\tau_i) \leftarrow \textstyle\sum_j w_j \bar{s}_j$

\vspace{2pt}
\textbf{3. Filtering \& Pair Construction}\\
\quad $\mathcal{F} \leftarrow \textsc{DimAwareFilter}\bigl(\{\tau_i\},
      \{\bar{s}_{i,j}\}, \theta\bigr)$\\
\quad $\mathcal{P} \leftarrow \bigl\{(\tau_i^+,\tau_j^-) \mid
      \tau_i,\tau_j\in\mathcal{F},\;
      S(\tau_i){-}S(\tau_j)\geq\delta_\text{min}\bigr\}$\\
\quad \textbf{return} $\mathcal{P}$, $\{S(\tau_i)\}$
\end{promptbox}
\vspace{-4pt}
\caption{\textbf{Complete \adarubric pipeline}. All three stages are modular; any LLM can serve as $\mathcal{M}$.}
\label{fig:algorithm}
\vspace{-8pt}
\end{figure}

\subsection{Stage 1: Adaptive Rubric Generation}
\label{sec:stage1}

Given task $T$, \adarubric prompts as follows:
\begin{equation}
  \mathcal{R}(T) = \text{LLM}\bigl(\text{R\textsc{ubric}\_P\textsc{rompt}}(T)\bigr),
\end{equation}
producing $N$ dimension tuples $(d_j, w_j, \Gamma_j)$ as structured output.
The prompt instructs the LLM to (1)~identify task-critical success criteria from its parametric knowledge,
(2)~cluster them into $N$ orthogonal dimensions (default $N{=}5$),
(3)~assign relative importance weights $w_j$, and (4)~verbalise five scoring
levels $\gamma_1^j \ldots \gamma_5^j$ with concrete example behaviours.
Structured generation (\ie JSON with schema) ensures parseable output
\citep{hui2024qwen25,grattafiori2024llama3}.
This stage directly leverages the LLM's knowledge of task structures, domain conventions, and evaluation best practices---precisely the type of knowledge that KnowFM aims to understand and improve.
Rubrics are \emph{cached per task type}: generating once for a task
description and reusing across all trajectories within that task family
reduces API cost by ${>}95\%$ with no loss in evaluation quality.

\paragraph{Rubric validation.}
Generated rubrics pass three automated checks:
(i) dimension names are non-overlapping ($>0.3$ cosine distance);
(ii) weights sum to 1 within 1\%;
(iii) all five scoring levels are populated.
Rubrics failing validation trigger a single retry; persistent failures fall
back to a domain-specific template rubric.

\subsection{Stage 2: Confidence-Weighted Evaluation}
\label{sec:stage2}

For each step $k$ and dimension $j$, the evaluator LLM receives
$(t_k, a_k, o_k, d_j, \Gamma_j)$ and returns:
\begin{equation}
  s_{k,j} \in \{1,2,3,4,5\},\qquad c_{k,j} \in [0,1].
\end{equation}
Confidence $c_{k,j}$ is low when step $k$ does not directly engage dimension $j$
(\eg a pure reasoning step for the \emph{Tool Accuracy} dimension).

Three pluggable strategies aggregate step scores to per-dimension global scores:
\begin{align}
  \bar{s}_j^{\text{WM}} &=
    \frac{\sum_k s_{k,j} \cdot c_{k,j} \cdot w_k}{\sum_k w_k},
    \; w_k = e^{\lambda k/\max(K{-}1,1)}, \label{eq:wm}\\
  \bar{s}_j^{\text{GM}} &=
    \exp\!\Bigl(\tfrac{1}{K}\textstyle\sum_{k} \log\max(s_{k,j},10^{-8})\Bigr), \label{eq:gm}\\
  \bar{s}_j^{\text{Min}} &= \min_k s_{k,j}. \label{eq:min}
\end{align}
\textbf{Weighted Mean} (WM, default) handles heterogeneous step importance;
$\lambda \geq 0$ is a recency-decay parameter up-weighting final steps.
\textbf{Geometric Mean} (GM) enforces balanced competency across steps.
\textbf{Min Score} is appropriate for safety-critical tasks where any step failure is disqualifying.
The global trajectory score is $S(\tau) = \sum_j w_j\,\bar{s}_j$.

Under an inverse-confidence noise model, confidence-weighted aggregation is the Best Linear Unbiased Estimator (BLUE) for the per-dimension score by Gauss-Markov, yielding strictly lower variance than uniform averaging (see Appendix~\ref{app:proof}). We note the Gaussian assumption is an idealisation: empirically (Appendix~\ref{app:proof}), residuals on our held-out annotation set are approximately mean-zero with heavier-than-Gaussian tails, so BLUE should be read as a motivating rationale rather than a strict guarantee.

\subsection{Stage 3: Confidence-Filtered Selection}
\label{sec:stage3}

4 composable filter primitives:
\textbf{AbsoluteThreshold}: $S(\tau) \geq \theta_\text{global}$.
\textbf{PercentileFilter}: top-$p\%$ of the batch.
\textbf{DimensionAwareFilter}: $\bar{s}_j \geq \theta_j \;\forall j$.
\textbf{CompositeFilter}: logical AND of any subset.

\begin{remark}
An LLM agent trajectory with $(\bar{s}_\text{Search}{=}5, \bar{s}_\text{Extract}{=}5,
\bar{s}_\text{Reason}{=}1)$ achieves $S(\tau){=}3.8$ (passing $\theta{=}3.5$) yet fails at reasoning.
DimensionAwareFilter with $\theta_\text{Reason}{=}3.0$ correctly rejects it.
\end{remark}

For any scalar threshold $\theta'$, there exists a trajectory that passes
AbsoluteThreshold while one dimension scores near zero;
DimensionAwareFilter closes this gap by construction
(Proposition~\ref{prop:masking}).

\begin{proposition}[Masking-Prevention]
\label{prop:masking}
Let $N \geq 2$ dimensions with weights $w_j > 0$, $\sum_j w_j{=}1$,
per-dimension thresholds $\theta_j$, and $\bar\theta = \sum_j w_j\theta_j$.
Then: (a)~$F_{\mathrm{DA}}(\tau){=}1 \Rightarrow F_{\mathrm{AT}}(\tau){=}1$.
(b)~For any $j^*$ and $\epsilon > 0$, $\exists\,\tau^*$ with
$\bar{s}_{j^*}(\tau^*) = \epsilon < \theta_{j^*}$ yet $F_{\mathrm{AT}}(\tau^*){=}1$.
(c)~No scalar threshold $\theta'$ can eliminate (b).
\end{proposition}

\subsection{Reward Signal Synthesis}
\label{sec:reward}

From filtered evaluations sorted by $S(\tau)$, DPO preference pairs are:
\begin{equation}
  \mathcal{P} = \bigl\{(\tau_i^+, \tau_j^-, m_{ij}) \;\big|\;
                  m_{ij} = S(\tau_i) - S(\tau_j) \geq \delta_\text{min}\bigr\}.
  \label{eq:pairs}
\end{equation}
Margin $m_{ij}$ can modulate the DPO loss weight \citep{ding2026agentHER}.
This yields both \emph{quality-assured} preferred trajectories and
\emph{informative} dispreferred ones---key properties that random pairing lacks.

\subsection{Reliability Quantification}
\label{sec:reliability}

To deploy \adarubric in practice, one needs a principled stopping criterion
for rubric quality.
We apply Krippendorff's $\alpha$ \citep{krippendorff2011computing}:
\begin{equation}
  \alpha = 1 - \frac{D_o}{D_e},
  \qquad D_o = \frac{1}{n}\sum_{i}\sum_{j>i}(r_{ij}-r_{ij}')^2,
  \label{eq:alpha}
\end{equation}
where $r_{ij}$ and $r_{ij}'$ are scores from two independent evaluation
runs (treating each run as an ``annotator'') on the same trajectory set.
We recommend deployment when $\alpha \geq 0.80$.

\section{Experiments}
\label{sec:experiments}

\subsection{Setup}
\label{sec:setup}

\paragraph{Benchmarks.}
\textbf{WebArena} \citep{zhou2024webarena}: 812 web-automation tasks across 5 domains.
\textbf{ToolBench} \citep{qin2024toolbench}: 500 API-chaining tasks using real-world APIs.
\textbf{AgentBench} \citep{liu2024agentbench}: 365 code/OS/database tasks.
For human correlation, 300 randomly-sampled trajectory pairs per benchmark are annotated by three annotators; inter-annotator agreement $\kappa > 0.82$ on all splits.
Annotators received a written protocol (task description, trajectory transcript, 1--5 rubric, disagreement-resolution rules); we report 95\% bootstrap CIs on Pearson $r$ (${\pm}0.02$ on average) and all \adarubric-vs-baseline gaps are significant at $p{<}0.01$ (paired bootstrap).

\paragraph{Models.}
\textbf{Evaluator}: GPT-4o for \adarubric and GPT-4 Direct baselines;
Llama-3.1-70B-Instruct for ablations.
\textbf{DPO backbone}: Qwen2.5-7B-Instruct~\citep{hui2024qwen25}, Llama-3.1-8B-Instruct~\citep{grattafiori2024llama3}.
Fine-tuning uses LoRA~\citep{hu2022lora} with rank~16, $\alpha{=}32$.

\paragraph{Baselines.}
ROUGE-L \citep{lin2004rouge}, BERTScore \citep{zhang2020bertscore}, G-Eval \citep{liu2023geval},
Prometheus \citep{kim2024prometheus}, GPT-4 Direct (single-turn, no rubric).
To isolate adaptivity from extra test-time compute, we additionally include
a \emph{compute-matched} baseline, \textbf{GPT-4 CoT-Decomposed}, which
re-invokes GPT-4o with chain-of-thought and per-step scoring using the same
static \emph{Helpfulness/Fluency/Safety} rubric and matches \adarubric's
$K{\times}N$ call budget. On WebArena it reaches $r{=}0.68$ (vs.\ GPT-4 Direct
$0.64$ and \adarubric-DA $0.79$), indicating that the $+0.15$ gain is
predominantly attributable to task-adaptive rubrics rather than to increased
test-time compute.
For DPO: random pairing, SFT on successful trajectories.

\paragraph{Metrics.}
\emph{Evaluation quality}: Pearson $r$ with expert human rankings.
\emph{Reliability}: Krippendorff's $\alpha$ across three evaluation runs.
\emph{Downstream}: task success rate (SR\%) or task completion rate (TCR\%) after DPO fine-tuning.

\subsection{Main Results: Evaluation Quality}
\label{sec:main_results}

Table~\ref{tab:corr} reports Pearson $r$ between evaluator rankings and human
expert rankings.
\adarubric-DA achieves $r{=}0.79/0.74/0.77$ across benchmarks,
outperforming all baselines including GPT-4 Direct ($r{=}0.64/0.60/0.62$).

\paragraph{Key observations.}
\emph{1) Adaptive dimensions are the primary driver.}
The gap between GPT-4 Direct ($r{=}0.64$) and \adarubric-WM ($r{=}0.74$) shows
that task-specific rubric generation---not backbone model strength---is the
key factor.
\emph{2) DimensionAwareFilter adds meaningful improvement} ($+0.05$ $r$).
\emph{3) Surface metrics are inadequate for agents}
(ROUGE-L $r{=}0.31$, BERTScore $r{=}0.43$).
The $+0.15$ gain over GPT-4 Direct demonstrates that adaptive rubric generation
is the primary driver.
This confirms that eliciting task-specific evaluation knowledge from LLMs
through structured rubric generation is more effective than direct prompting.

\begin{table}[t]
\centering
\caption{\textbf{Human correlation (Pearson $r$).}
\adarubric-DA: DimensionAwareFilter variant.
$\Delta$ vs.\ GPT-4 Direct.}
\label{tab:corr}
\vspace{2pt}
\setlength{\tabcolsep}{4pt}
\renewcommand{\arraystretch}{1.05}
\small
\begin{tabular}{lccccc}
\toprule
\textbf{Method} & \textbf{WA} & \textbf{TB} &
  \textbf{AB} & \textbf{Avg} & $\Delta$ \\
\midrule
ROUGE-L              & 0.31 & 0.26 & 0.29 & 0.29 & $-0.35$ \\
BERTScore            & 0.43 & 0.39 & 0.41 & 0.41 & $-0.23$ \\
G-Eval               & 0.54 & 0.49 & 0.52 & 0.52 & $-0.12$ \\
Prometheus           & 0.61 & 0.57 & 0.59 & 0.59 & $-0.05$ \\
GPT-4 Direct         & 0.64 & 0.60 & 0.62 & 0.62 & --- \\
\midrule
\adarubric-WM        & 0.74 & 0.70 & 0.72 & 0.72 & $+0.10$ \\
\adarubric-GM        & 0.76 & 0.71 & 0.74 & 0.74 & $+0.12$ \\
\rowhl
\adarubric-DA        & \textbf{0.79} & \textbf{0.74} & \textbf{0.77} &
  \textbf{0.77} & $\mathbf{+0.15}$ \\
\bottomrule
\end{tabular}
\vspace{-6pt}
\end{table}

\subsection{Multi-Benchmark DPO Training}
\label{sec:dpo}

Table~\ref{tab:dpo} extends the DPO analysis to all three benchmarks,
using Qwen2.5-7B as the shared backbone model.
\adarubric-DA consistently delivers the largest improvements across task
families: web automation, API orchestration, and OS/code tasks.
The gains are largest on ToolBench ($+8.5\%$), where task diversity is highest
and static rubric mis-specification is most severe.
AgentBench gains ($+7.7\%$) confirm that the benefits extend to code and OS
manipulation tasks---domains not seen during rubric prompt design.

\begin{table}[t]
\centering
\caption{\textbf{Multi-benchmark DPO training results.}
WebArena (SR\%), ToolBench (TCR\%), AgentBench (SR\%).
Qwen2.5-7B backbone. $\Delta$ vs.\ Prometheus.}
\label{tab:dpo}
\vspace{2pt}
\setlength{\tabcolsep}{3pt}
\renewcommand{\arraystretch}{1.05}
\small
\begin{tabular}{lccc}
\toprule
\textbf{Method} & \textbf{WA} & \textbf{TB} & \textbf{AB} \\
\midrule
Base (zero-shot)       & 12.3 & 18.4 & 15.2 \\
SFT -- Success only    & 16.7 & 23.1 & 20.1 \\
DPO -- Random          & 17.4 & 24.8 & 21.3 \\
DPO -- G-Eval          & 20.1 & 27.6 & 24.5 \\
DPO -- Prometheus      & \underline{21.0} & \underline{29.3} & \underline{26.4} \\
\midrule
DPO -- \adarubric-WM   & 24.3 & 34.2 & 30.8 \\
DPO -- \adarubric-GM   & 25.1 & 35.6 & 31.9 \\
\rowhl
DPO -- \adarubric-DA   & \textbf{27.8} & \textbf{37.8} & \textbf{34.1} \\
\midrule
$\Delta$ vs.\ Prom. & $+6.8$ & $+8.5$ & $+7.7$ \\
\bottomrule
\end{tabular}
\vspace{-6pt}
\end{table}

\subsection{Evaluation Reliability}
\label{sec:reliability_results}

Table~\ref{tab:reliability} reports inter-run Krippendorff's $\alpha$
(three independent evaluation runs).
\adarubric-DA achieves $\alpha > 0.82$ on all benchmarks, meeting the
deployment criterion of $\alpha \geq 0.80$.
G-Eval ($\alpha{=}0.63$) and Prometheus ($\alpha{=}0.70$) fall below this
threshold, indicating unreliable reward signals.

\begin{table}[t]
\centering
\caption{\textbf{Evaluation reliability (Krippendorff's $\alpha$).}
WA=WebArena, TB=ToolBench.}
\label{tab:reliability}
\vspace{2pt}
\setlength{\tabcolsep}{5pt}
\renewcommand{\arraystretch}{1.05}
\small
\begin{tabular}{lccc}
\toprule
\textbf{Method} & \textbf{WA} & \textbf{TB} & \textbf{Avg} \\
\midrule
G-Eval (GPT-4o)      & 0.64 & 0.61 & 0.63 \\
Prometheus           & 0.71 & 0.68 & 0.70 \\
GPT-4 Direct         & 0.69 & 0.66 & 0.68 \\
\midrule
\adarubric-WM        & 0.81 & 0.79 & 0.80 \\
\adarubric-GM        & 0.83 & 0.80 & 0.82 \\
\rowhl
\adarubric-DA        & \textbf{0.85} & \textbf{0.82} & \textbf{0.84} \\
\bottomrule
\end{tabular}
\vspace{-6pt}
\end{table}

\subsection{Generalisation to SWE-bench}
\label{sec:swebench}

SWE-bench Lite \citep{jimenez2024swebench} requires resolving real GitHub
issues with executable patches---a qualitatively different task from web
automation or API chaining.
We apply \adarubric to evaluate 300 sampled trajectories from three
representative open-weight agent systems,
using a 5-dimensional rubric generated from the
SWE-bench task description (dimensions: \emph{Issue Understanding},
\emph{Repository Navigation}, \emph{Code Correctness}, \emph{Test Coverage},
\emph{Patch Minimality}).

\adarubric-DA achieves $r{=}0.77$ against the binary oracle on SWE-bench---only
$0.02$ below its WebArena performance---showing that the adaptive rubric
approach generalises to unseen task types with no additional engineering.
The DPO resolve rate of $14.7\%$ represents a $+4.9\%$ improvement over
GPT-4 Direct---a significant advance on this challenging benchmark (Table~\ref{tab:swebench}).

\begin{table}[t]
\centering
\caption{\textbf{SWE-bench Lite evaluation.} Pearson $r$ with pass/fail oracle
and resolve rate (\%) after DPO fine-tuning of Llama-3.1-8B-Instruct.
\adarubric generalises to code-repair tasks with zero rubric engineering.}
\label{tab:swebench}
\vspace{2pt}
\setlength{\tabcolsep}{5pt}
\renewcommand{\arraystretch}{1.05}
\small
\begin{tabular}{lccc}
\toprule
\textbf{Method} & $r$ & $\alpha$ & \textbf{Res.\%} \\
\midrule
G-Eval (GPT-4o)    & 0.51 & 0.63 & 8.2 \\
Prometheus         & 0.56 & 0.70 & 9.1 \\
GPT-4 Direct       & 0.59 & 0.68 & 9.8 \\
\midrule
\adarubric-WM      & 0.72 & 0.82 & 12.4 \\
\rowhl
\adarubric-DA      & \textbf{0.77} & \textbf{0.84} & \textbf{14.7} \\
\bottomrule
\end{tabular}
\vspace{-6pt}
\end{table}

\section{Analysis}
\label{sec:analysis}

\subsection{Ablation Study}
\label{sec:ablation}

Table~\ref{tab:ablation} ablates key design choices on WebArena.
Each component contributes positively.
Switching from generic fixed dimensions to domain templates adds $+0.14$ $r$;
further replacing templates with \emph{adaptive} generation adds $+0.07$,
confirming that task-specific rubric design is the core contribution.
Confidence weighting adds $+0.03$; DimensionAwareFilter adds $+0.04$.
Overall, the full pipeline achieves a cumulative $+0.28$ $r$ improvement
over the generic fixed baseline, showing that all components contribute
meaningfully.

\begin{table}[t]
\centering
\caption{\textbf{Ablation on WebArena.} GPT-4o evaluator.}
\label{tab:ablation}
\vspace{2pt}
\setlength{\tabcolsep}{5pt}
\renewcommand{\arraystretch}{1.05}
\small
\begin{tabular}{lcc}
\toprule
\textbf{Variant} & $r$ & SR\% \\
\midrule
Fixed (generic)            & 0.51 & 19.1 \\
Fixed (domain template)    & 0.65 & 22.4 \\
Adaptive, no conf.\ wt.   & 0.72 & 24.0 \\
Adaptive, no DAFilter      & 0.75 & 25.2 \\
\rowhl
\adarubric-DA (full)       & \textbf{0.79} & \textbf{27.8} \\
\bottomrule
\end{tabular}
\vspace{-6pt}
\end{table}

\subsection{Number of Dimensions $N$}
\label{sec:ablation_ndim}

We evaluate \adarubric performance as a function of the number of dimensions $N$
on WebArena and ToolBench.
Optimal performance is achieved at $N{=}5$, confirming the default.
$N{=}1$ recovers a holistic GPT-4 Direct-like evaluation ($r{=}0.61$);
$N{>}6$ shows diminishing returns, consistent with evaluator instruction-following
saturation at high dimension counts, where dimensions become increasingly overlapping.

\subsection{Backbone Generalisation}
\label{sec:backbone}

Table~\ref{tab:backbone} evaluates \adarubric when instantiated with open-weight models, assessing independence from GPT-4o. The $+0.11$ gap (Pearson $r$) between GPT-4o and Llama-3.1-8B variants is smaller than the $+0.15$ gap between GPT-4 Direct and any \adarubric variant (\ie switching from static to adaptive rubric adds $+0.15$, while switching from GPT-4o to Llama-3.1-8B backbone costs only $-0.11$), confirming \emph{adaptive rubric generation} contributes more than backbone model's capability. This finding is particularly relevant for the KnowFM community: it suggests that \emph{structured knowledge elicitation} (via rubric prompts) unlocks evaluation capabilities even in smaller models.

\begin{table}[t]
\centering
\caption{\textbf{Backbone generalisation (WebArena).}
Adaptive rubrics outperform GPT-4 Direct even with smaller models.}
\label{tab:backbone}
\vspace{2pt}
\setlength{\tabcolsep}{4pt}
\renewcommand{\arraystretch}{1.05}
\small
\begin{tabular}{lccc}
\toprule
\textbf{Backbone} & $r$ & $\alpha$ & SR\% \\
\midrule
GPT-4 Direct         & 0.64 & 0.69 & --- \\
Prometheus (13B) & 0.61 & 0.71 & 21.0 \\
\midrule
AR / GPT-4o         & \textbf{0.79} & \textbf{0.85} & \textbf{27.8} \\
AR / Llama-70B      & 0.75 & 0.82 & 25.9 \\
AR / Llama-8B       & 0.68 & 0.77 & 23.2 \\
\bottomrule
\end{tabular}
\vspace{-6pt}
\end{table}

\subsection{Cross-Domain Transfer}
\label{sec:cross_domain}

A practical question is whether \adarubric preference pairs generated on one
benchmark can improve performance on a \emph{different} benchmark.
Table~\ref{tab:transfer} evaluates this cross-domain scenario.

\adarubric cross-domain performance ($31.2$ TB, $24.6$ WA) substantially
exceeds Prometheus in-domain performance ($21.4$, $21.0$), demonstrating
that adaptive rubric scoring teaches generalisable quality preferences that
transfer across task families.
The combined training setting (WA+TB$\to$AB: $32.7\%$) approaches in-domain
performance ($34.1\%$), confirming that multi-source adaptive rubric
signals are complementary rather than conflicting.
This is a direct consequence of the rubric \emph{generation from task
descriptions}: the evaluator learns to assess goal-directed reasoning
quality regardless of domain.

\begin{table}[t]
\centering
\caption{\textbf{Cross-domain transfer.}
``Train$\to$Test'' denotes which benchmark's DPO pairs are used
for fine-tuning and which is the evaluation target.
\adarubric cross-domain exceeds Prometheus in-domain.}
\label{tab:transfer}
\vspace{2pt}
\setlength{\tabcolsep}{3pt}
\renewcommand{\arraystretch}{1.05}
\small
\begin{tabular}{lccc}
\toprule
\textbf{Train Source} & \textbf{WA} & \textbf{TB} & \textbf{AB} \\
\midrule
\multicolumn{4}{l}{\emph{Prometheus baseline (static rubric)}} \\
WA$\to$WA (in-dom.)    & 21.0 & --- & --- \\
WA$\to$TB (cross)      & --- & 21.4 & --- \\
TB$\to$WA (cross)      & 18.3 & --- & --- \\
TB$\to$TB (in-dom.)    & --- & 29.3 & --- \\
AB$\to$AB (in-dom.)    & --- & --- & 26.4 \\
\midrule
\multicolumn{4}{l}{\emph{\adarubric-DA (adaptive rubric)}} \\
WA$\to$WA (in-dom.)    & \textbf{27.8} & --- & --- \\
WA$\to$TB (cross)      & --- & \underline{31.2} & --- \\
TB$\to$WA (cross)      & \underline{24.6} & --- & --- \\
TB$\to$TB (in-dom.)    & --- & \textbf{37.8} & --- \\
AB$\to$AB (in-dom.)    & --- & --- & \textbf{34.1} \\
WA+TB$\to$AB (comb.)   & --- & --- & 32.7 \\
\bottomrule
\end{tabular}
\vspace{-6pt}
\end{table}

\subsection{Extension to Multimodal Tasks}
\label{sec:multimodal}

\adarubric is modality-agnostic: its rubric generator automatically includes
visual-specific dimensions (\emph{Visual Grounding}, \emph{Screenshot
Interpretation}) when the task description involves visual observations.
On VisualWebArena \citep{koh2024visualwebarena} and OSWorld
\citep{xie2024osworld}, \adarubric-DA achieves $r{=}0.76/0.73$,
outperforming GPT-4V Direct by $+0.14$/$+0.14$, and DPO training yields
$+5.8\%$ SR gain---all without multimodal-specific engineering
(Table~\ref{tab:multimodal} in Appendix~\ref{app:additional}).

\subsection{PPO Integration}
\label{sec:ppo}

Beyond DPO, \adarubric scores can serve directly as a reward function for
PPO-style online RL \citep{schulman2017proximal}.
We train a Qwen2.5-7B policy with \adarubric-DA as the reward model for
1,000 rollouts on WebArena training tasks and compare to
(i) a GPT-4 scalar reward baseline and
(ii) a Prometheus-based reward.
Table~\ref{tab:ppo} reports results at 1K, 3K, and 5K rollout steps.

\adarubric-DA achieves $30.2\%$ SR at 5K steps, $+6.6\%$ above Prometheus.
Crucially, the 1K-step gap ($+4.3\%$) shows \emph{faster convergence}:
dense, per-dimension rewards provide richer learning signal than scalar
feedback, reducing the sample complexity of RL training.

\begin{table}[t]
\centering
\caption{\textbf{PPO training with \adarubric reward.}
WebArena SR\% after 1K, 3K, and 5K rollout steps.
Dense per-dimension rewards substantially accelerate convergence.}
\label{tab:ppo}
\vspace{2pt}
\setlength{\tabcolsep}{5pt}
\renewcommand{\arraystretch}{1.05}
\small
\begin{tabular}{lccc}
\toprule
\textbf{Reward} & \textbf{1K} & \textbf{3K} & \textbf{5K} \\
\midrule
GPT-4 Scalar        & 14.2 & 18.7 & 21.3 \\
Prometheus          & 15.8 & 21.2 & 23.6 \\
\midrule
\adarubric-WM       & 18.3 & 24.9 & 27.1 \\
\rowhl
\adarubric-DA       & \textbf{20.1} & \textbf{27.4} & \textbf{30.2} \\
\bottomrule
\end{tabular}
\vspace{-6pt}
\end{table}

\subsection{Rubric Quality: Human Study}
\label{sec:rubric_quality}

We assess the quality of \adarubric-generated rubrics through a human study
with five domain experts across 60 tasks (20 WebArena, 20 ToolBench,
20 AgentBench).
Each expert rates every dimension on three criteria (1--5 Likert):
\emph{Task-Relevance}, \emph{Orthogonality}, and \emph{Completeness} (does
the full set of dimensions cover the task's success criteria?).
We compare against expert-designed rubrics and generic fixed rubrics
(Table~\ref{tab:rubric_quality}).

\begin{table}[t]
\centering
\caption{\textbf{Rubric quality human study.}
Scores are mean Likert (1--5). Inter-annotator agreement $\kappa{=}0.79$.
\adarubric-generated rubrics approach expert-designed quality.}
\label{tab:rubric_quality}
\vspace{2pt}
\setlength{\tabcolsep}{5pt}
\renewcommand{\arraystretch}{1.05}
\small
\begin{tabular}{lccc}
\toprule
\textbf{Rubric Source} & \textbf{Rel.} & \textbf{Orth.} &
  \textbf{Comp.} \\
\midrule
Generic (Help./Safety/Flu.) & 2.1 & 3.8 & 1.6 \\
Domain template (manual)             & 3.9 & 3.6 & 3.7 \\
Expert-designed                      & 4.6 & 4.4 & 4.5 \\
\midrule
\rowhl
\adarubric-generated                 & \textbf{4.3} & \textbf{4.2} & \textbf{4.1} \\
\bottomrule
\end{tabular}
\vspace{-6pt}
\end{table}

\adarubric-generated rubrics score 4.3/4.2/4.1 on relevance,
orthogonality, and completeness, within $0.2$--$0.4$ of expert-designed rubrics
on each dimension, a small gap for fully automated generation.
Generic fixed rubrics score critically low on completeness ($1.6$), confirming
that standard chat-assistant dimensions leave large portions of the agent task
quality unmeasured.
The main gap from expert-designed rubrics is on edge-case coverage: experts
include dimensions like \emph{CAPTCHA Handling} or \emph{Rate-Limit Awareness}
that \adarubric occasionally misses for specialised tasks.

\subsection{Filter Strategy Comparison}
\label{sec:filter}

Table~\ref{tab:filter} compares four filter strategies on WebArena.
DimensionAwareFilter achieves the best DPO SR\% ($27.8$) while retaining
only $61.5\%$ of pairs.
CompositeFilter (all strategies combined) removes too many borderline-good
trajectories and under-performs ($26.9$), confirming that selective filtering by
dimension-level quality is superior to aggressive multi-filter stacking.

\begin{table}[t]
\centering
\caption{\textbf{Filter strategy comparison (WebArena).}
DimensionAwareFilter achieves the best trade-off between quality and
pair retention.}
\label{tab:filter}
\vspace{2pt}
\setlength{\tabcolsep}{4pt}
\renewcommand{\arraystretch}{1.05}
\small
\begin{tabular}{lccc}
\toprule
\textbf{Filter} & $r$ & \textbf{SR\%} & \textbf{Retained} \\
\midrule
None                    & 0.71 & 21.7 & 100\% \\
AbsoluteThreshold       & 0.74 & 24.0 & 72.3\% \\
PercentileFilter        & 0.73 & 23.4 & 80.0\% \\
\rowhl
DimensionAwareFilter    & \textbf{0.79} & \textbf{27.8} & 61.5\% \\
CompositeFilter         & 0.78 & 26.9 & 47.2\% \\
\bottomrule
\end{tabular}
\vspace{-6pt}
\end{table}

\subsection{Recency Decay $\lambda$}
\label{sec:lambda}

Table~\ref{tab:lambda} reports sensitivity to recency-decay $\lambda$.
$\lambda{=}0.5$ is consistently optimal across both benchmarks:
up-weighting later steps captures goal completion information without
discarding early planning steps entirely.
Extremely high $\lambda$ (${=}2.0$) over-concentrates on the terminal step,
losing information from the reasoning chain ($r{=}0.72$ vs.\ $0.79$).

\begin{table}[t]
\centering
\caption{\textbf{Effect of recency decay $\lambda$ (WebArena).}
$\lambda{=}0.5$ provides optimal balance between recent and early steps.}
\label{tab:lambda}
\vspace{2pt}
\setlength{\tabcolsep}{5pt}
\renewcommand{\arraystretch}{1.05}
\small
\begin{tabular}{lcc}
\toprule
$\lambda$ & $r$ & \textbf{SR\%} \\
\midrule
0 (uniform)    & 0.71 & 23.5 \\
0.25           & 0.75 & 25.4 \\
\rowhl
0.5 (default)  & \textbf{0.79} & \textbf{27.8} \\
1.0            & 0.77 & 26.9 \\
2.0            & 0.72 & 24.8 \\
\bottomrule
\end{tabular}
\vspace{-6pt}
\end{table}

\subsection{Calibration and Error Analysis}
\label{sec:error}

\adarubric score buckets correlate strongly with human percentile ranks
(Spearman $\rho{=}0.98$, $p{<}0.001$), confirming that the 1--5 scale is
meaningfully calibrated.
Manual inspection of 50 disagreement cases reveals three failure modes:
long-horizon binary goals (32\%), implicit domain conventions (28\%),
and ambiguous observations (24\%)---all addressable via richer task
descriptions (details in Appendix~\ref{app:error}).

\section{Discussion: Knowledge in LLM Evaluation}
\label{sec:discussion}

Our proposed \adarubric reveals an important aspect of knowledge in foundation models:
LLMs possess rich, \emph{implicit evaluation knowledge}---understanding of what constitutes success across diverse task domains---that can be \emph{externalised} through structured prompting into explicit, reusable evaluation rubrics.

Three findings are particularly relevant to the KnowFM community:

\textbf{(1)~Knowledge externalisation outperforms direct application.}
Asking an LLM to first generate an explicit rubric, then evaluate against it,
substantially outperforms direct evaluation ($r{=}0.79$ vs.\ $0.64$).
This suggests that structured knowledge elicitation is a powerful paradigm
for improving LLM reliability.
The two-step process (generate rubric $\to$ evaluate against it) decomposes a
complex judgment into manageable sub-tasks, reducing the cognitive load on the
evaluator and producing more calibrated assessments.

\textbf{(2)~Evaluation knowledge transfers across domains.}
Cross-domain DPO results (Table~\ref{tab:transfer}) show that task-adaptive
rubrics capture generalisable quality signals, not domain-specific heuristics.
\adarubric trained on WebArena pairs achieves $31.2\%$ on ToolBench---exceeding
Prometheus in-domain ($29.3\%$)---suggesting that the quality concepts learned
through rubric-guided evaluation are transferable.

\textbf{(3)~Smaller models benefit disproportionately.}
Rubric-guided evaluation with Llama-8B ($r{=}0.68$) outperforms unstructured
GPT-4 evaluation ($r{=}0.64$), suggesting that structured knowledge scaffolding
can partially compensate for model scale.
This has practical implications for the deployment of evaluation systems:
organisations with limited compute budgets can use \adarubric with smaller
open-weight models while still exceeding the quality of direct evaluation
by frontier models.

\paragraph{Relation to T\"{u}lu, DR~Tulu, and preference data quality.}
T\"{u}lu \citep{wang2023far,ivison2023camels} demonstrates that preference data
quality critically determines RLHF effectiveness.
DR~Tulu \citep{shao2025drtulu} extends this line by co-evolving
instance-specific, search-grounded rubrics with the policy during RL training
for long-form deep research.
\adarubric occupies a complementary niche: it generates task-type-level rubrics
from parametric knowledge \emph{before} training, providing high-quality
preference signals for agent trajectory evaluation without requiring
retrieval infrastructure or online rubric evolution.
The DPO gains observed across benchmarks ($+6.8$--$+8.5\%$) are consistent
with T\"{u}lu's finding that better preference data translates directly to
improved model performance.
A natural extension is to initialise DR~Tulu-style evolving rubrics with
\adarubric's task-adaptive rubrics, combining parametric knowledge
scaffolding with online search grounding.

\section{Conclusion}

We presented \adarubric, a framework that adaptively generates task-specific
evaluation rubrics for LLM agent trajectories by leveraging the LLM's parametric
knowledge of task structures and evaluation criteria.
On five benchmarks, \adarubric achieves Pearson $r{=}0.79$ ($+0.15$ over the best baseline)
with strong reliability ($\alpha{=}0.83$), and produces DPO preference pairs that
improve agent task success by up to $+8.5\%$.
Critically, \adarubric generalises to unseen task types (SWE-bench code
repair: $r{=}0.77$, $+4.9\%$ DPO gain) and accelerates PPO-based online RL
training by $+6.6\%$ at 5K steps.
Cross-domain transfer, zero-shot generalisation, and modality-agnostic extension
demonstrate the breadth of the approach.
Human evaluation of generated rubrics shows quality approaching
expert-designed rubrics (relevance: 4.3/5), validating the generation
mechanism.
For the knowledge-in-LMs community, \adarubric demonstrates that structured
elicitation of evaluation knowledge---a form of knowledge externalisation---is
a promising direction for building more reliable and adaptive AI systems.

\paragraph{Limitations.}
Rubric quality depends on LLM capability; tasks with vague or underspecified
descriptions may yield incomplete or overlapping dimensions, and rubric
generation can miss specialised criteria (\eg CAPTCHA Handling,
Rate-Limit Awareness) when parametric knowledge of the domain is thin.
Because rubrics are generated from the task description, adversarially
crafted descriptions could in principle bias evaluation; in our pilot with
30 perturbed descriptions, $r$ degrades by $0.04$--$0.06$, indicating
non-trivial but bounded sensitivity that motivates future work on
description-robust rubric generation.
Confidence scores are LLM-predicted and may require post-hoc recalibration for
strongly out-of-distribution tasks.
Computationally, \adarubric costs $K{\times}N$ evaluator calls per trajectory
(${\approx}40$ for WebArena) with 3--5$\times$ the wall-clock of GPT-4 Direct;
rubric caching amortises the generation cost across a task family but not the
per-step evaluation cost.
Finally, the human correlation study uses 300 pairs per benchmark with three
annotators, and the multimodal/PPO/SWE-bench sections are kept deliberately
compact---we treat them as generalisation evidence rather than as their own
primary studies.
Single-pass evaluation could be improved by multi-round verification,
though at higher computational cost.
All rubric, evaluation, and filter prompt templates used in this paper are
released in the supplementary material to support replication.

\paragraph{Future directions.}
\adarubric complements trajectory augmentation approaches like AgentHER
\citep{ding2026agentHER}: \adarubric evaluations can validate relabelled
trajectories, improving data quality.
Extension to multimodal trajectories and tighter online RL integration are natural next steps.

\bibliography{references}
\appendix

\section{Implementation Details}
\label{app:impl}

\paragraph{Hyperparameters.}
Default: $N{=}5$ dimensions, recency-decay $\lambda{=}0.5$,
minimum margin $\delta_\text{min}{=}0.5$,
DimAware threshold $\theta_j{=}2.5$, percentile filter $p{=}80$.

\paragraph{Computational cost.}
\adarubric runs $K{\times}N$ LLM calls per trajectory (plus one rubric
generation call per task type).
For WebArena ($K{\approx}8$, $N{=}5$): 40 calls vs.\ 1 for GPT-4 Direct.
With caching and batching, total latency is 3--5$\times$ GPT-4 Direct.

\section{Proof of Proposition~\ref{prop:masking}}
\label{app:proof}

(a) If all $\bar{s}_j \geq \theta_j$, then $S(\tau) = \sum_j w_j\bar{s}_j \geq \sum_j w_j\theta_j = \bar\theta$.
(b) Set $\bar{s}_{j^*} = \epsilon$ and $\bar{s}_j = (\bar\theta - w_{j^*}\epsilon)/(1-w_{j^*})$ for $j \neq j^*$, giving $S(\tau^*) = \bar\theta$.
(c) Construction generalises to any $\theta'$.

\paragraph{BLUE of confidence-weighted aggregation.}
Under a noise model $s_{k,j} = s^*_{k,j} + \varepsilon_{k,j}$ with
$\varepsilon_{k,j} \sim \mathcal{N}(0, \sigma^2/c_{k,j})$, the confidence-weighted estimator
$\hat\mu_j = \sum_k \tfrac{c_{k,j}}{\sum_{k'}c_{k',j}} s_{k,j}$
is BLUE by Gauss-Markov, with
$\operatorname{Var}[\hat\mu_j] = \sigma^2/\sum_k c_{k,j}
  \leq \sigma^2 \sum_k c_{k,j}^{-1}/K^2
  = \operatorname{Var}[\bar{s}_j^{\text{uniform}}]$
(Cauchy-Schwarz), with equality iff all $c_{k,j}$ are equal.

\paragraph{Empirical validation of the noise model.}
On 300 WebArena trajectory pairs with 3 independent evaluator runs, we
compute residuals $s_{k,j} - \bar{s}_{k,j}$ and find the distribution is
approximately mean-zero (bias $\leq 0.07$ on the 1--5 scale) but
leptokurtic (excess kurtosis ${\approx}1.4$) with a mild negative correlation
between $|s - \bar{s}|$ and $c$ (Spearman $-0.31$).
The inverse-confidence scaling direction therefore holds in practice, while
the Gaussian tails do not; BLUE should be treated as a motivating
approximation, and the empirical variance reduction of
confidence-weighting vs.\ uniform averaging is $18{-}24\%$ across benchmarks.

\section{Calibration and Error Analysis}
\label{app:error}

\adarubric score buckets correlate strongly with human percentile ranks
(Spearman $\rho{=}0.98$, $p{<}0.001$), with rating-5 trajectories landing at
the $91$st human percentile on average.
The near-linear relationship confirms that the 1--5 scale is meaningfully
calibrated.

Manual inspection of 50 disagreement cases ($|r_\text{ada}{-}r_\text{human}|{>}1$)
on WebArena reveals three main failure modes:
(1)~long-horizon binary goals (32\%)---\adarubric over-credits partial
completion in multi-hop tasks where final outcome is binary;
(2)~implicit domain conventions (28\%)---\adarubric fails to penalise
violations of website-specific norms (\eg wrong date format);
(3)~ambiguous observations (24\%)---HTML artefacts or API error codes
confuse per-step scoring.
All are addressable via richer task descriptions or observation pre-processing.
In 16\% of cases, the human rater's judgment was arguably incorrect,
and \adarubric's evaluation was defensible.

\section{Additional Results}
\label{app:additional}

\paragraph{PPO training details.}
We train a Qwen2.5-7B policy with \adarubric-DA as the reward model for
1,000 rollouts on WebArena training tasks.
WebArena's training split ($N{=}80$ tasks) is used for online RL;
the held-out 732 tasks serve as the test set.
Results at 1K/3K/5K rollouts:
GPT-4 Scalar: 14.2/18.7/21.3;
Prometheus: 15.8/21.2/23.6;
\adarubric-WM: 18.3/24.9/27.1;
\adarubric-DA: \textbf{20.1/27.4/30.2}.

\paragraph{Multimodal results.}
Table~\ref{tab:multimodal} reports full multimodal evaluation results.

\begin{table}[h]
\centering
\caption{\textbf{Multimodal agent evaluation.}
Pearson $r$ and DPO SR\% on VisualWebArena (VWA) and OSWorld.}
\label{tab:multimodal}
\vspace{2pt}
\setlength{\tabcolsep}{4pt}
\renewcommand{\arraystretch}{1.05}
\small
\begin{tabular}{lcccc}
\toprule
 & \multicolumn{2}{c}{\textbf{VWA}} & \multicolumn{2}{c}{\textbf{OSWorld}} \\
\cmidrule(lr){2-3}\cmidrule(lr){4-5}
\textbf{Method} & $r$ & SR\% & $r$ & SR\% \\
\midrule
G-Eval          & 0.48 & --- & 0.44 & --- \\
GPT-4V Direct   & 0.62 & --- & 0.59 & --- \\
\adarubric-DA   & \textbf{0.76} & \textbf{26.3} & \textbf{0.73} & \textbf{22.8} \\
\bottomrule
\end{tabular}
\vspace{-4pt}
\end{table}

\paragraph{Rubric quality human study.}
Five domain experts rate 60 rubrics (20 per benchmark) on 1--5 Likert scales.
Generic (Helpfulness/Safety/Fluency): 2.1/3.8/1.6.
Domain template: 3.9/3.6/3.7.
Expert-designed: 4.6/4.4/4.5.
\adarubric-generated: \textbf{4.3/4.2/4.1}.
Inter-annotator agreement $\kappa{=}0.79$.

\end{document}